\documentclass{article}
\usepackage[preprint]{log_2025}				

\usepackage{algorithm}
\usepackage{algpseudocode}
\usepackage{adjustbox}

\usepackage[inline]{enumitem}
\usepackage{times}
\usepackage{latexsym}
\usepackage{booktabs} 
\usepackage{amsmath} 
\usepackage[T1]{fontenc}
\usepackage{multirow}
\usepackage{booktabs}
\usepackage{graphicx}
\usepackage{array}
\usepackage{caption}
\usepackage{pgfplots}
\pgfplotsset{compat=1.17}
\usepackage[utf8]{inputenc}
\usepackage[caption=false]{subfig}
\usepackage{wrapfig}

\usepackage{booktabs}						
\usepackage{multirow}						
\usepackage{amsfonts}						
\usepackage{graphicx}						
\usepackage{duckuments}						

\usepackage[numbers,compress,sort]{natbib}	

\title[SoG]{Synthesize-on-Graph: Knowledgeable Synthetic Data Generation for Continued Pre-training of Large Language Models}

\author{Shengjie Ma$^{1, 3}$\thanks{Both authors contributed equally to this research.}, Xuhui Jiang$^{1}$\footnotemark[1], Chengjin Xu$^{1}$\footnotemark[1] , \textbf{ Cehao Yang}$^{2}$, \textbf{Liyu Zhang}$^{2}$ \textbf{Jian Guo}$^{2}$\thanks{Corresponding author}\\
$^1$ DataArc Tech Ltd., Shenzhen, Guangdong, China \\
$^2$ IDEA Research, International Digital Economy Academy, Shenzhen, Guangdong, China \\
$^3$ Gaoling School of Artificial Intelligence, Renmin University of China, Beijing, China \\
}
\begin{document}

\maketitle
\begin{abstract}
Large Language Models (LLMs) have achieved remarkable success but remain data-inefficient, especially when learning from small, specialized corpora with limited and proprietary data. Existing synthetic data generation methods for continue pre-training focus on intra-document content and overlook cross-document knowledge associations, limiting content diversity and depth.
We propose Synthetic-on-Graph (SoG), a synthetic data generation framework that incorporates cross-document knowledge associations for efficient corpus expansion. SoG constructs a context graph by extracting entities and concepts from the original corpus, representing cross-document associations, and employing a graph walk strategy for knowledge-associated sampling. This enhances synthetic data diversity and coherence, enabling models to learn complex knowledge structures and handle rare knowledge.
To further improve the quality of synthetic data, we integrate two complementary strategies, Chain-of-Thought (CoT) and Contrastive Clarifying (CC), to enhance both reasoning capability and discriminative power. Extensive experiments demonstrate that SoG surpasses state-of-the-art (SOTA) methods on multi-hop and domain-specific question answering, while achieving competitive performance on long-context reading comprehension. These results highlight the superior generalization ability of SoG. Our work advances the paradigm of synthetic data generation and offers practical solutions for efficient knowledge acquisition in LLMs, particularly for downstream tasks and domains with limited training data.

\end{abstract}

\section{Introduction}
In recent years, Large Language Models (LLMs) have achieved groundbreaking advancements in the field of Natural Language Processing (NLP), demonstrating the ability to acquire knowledge from unstructured text and perform complex, knowledge-intensive tasks \citep{Brown2020}. These models have exhibited exceptional performance across various applications, including question-answering systems, machine translation, and conversational agents. This success is largely attributed to the next-word prediction objective \citep{Shannon1951} combined with vast amounts of internet data \citep{CommonCrawl2007}. However, despite these achievements, there remains a significant inefficiency in data utilization~\citep{wan2024efficientlargelanguagemodels}.

This data inefficiency becomes particularly pronounced when models need to learn from small-scale, high-value corpora. With the increasing demand for proprietary domain knowledge, models are required to efficiently acquire information from limited data sources. For instance, in specialized fields such as medicine, law, or specific technological domains, the available data is not only limited but often proprietary. In such cases, traditional large-scale pretraining methods are inapplicable due to the unavailability of sufficient training data \citep{wang2024knowledgemechanismslargelanguage}.

Moreover, recent studies have revealed limitations in the current pretraining paradigm. For example, models struggle when learning simple relations and require a large number of repeated instances to effectively learn facts \citep{Allen-ZhuLi2024}. These issues become more acute when dealing with long-tail data or rare knowledge, as such information appears with extremely low frequency in large-scale corpora \citep{wang2024knowledgemechanismslargelanguage}.

\begin{figure}[h]
\begin{center}
\includegraphics[width=1\columnwidth]{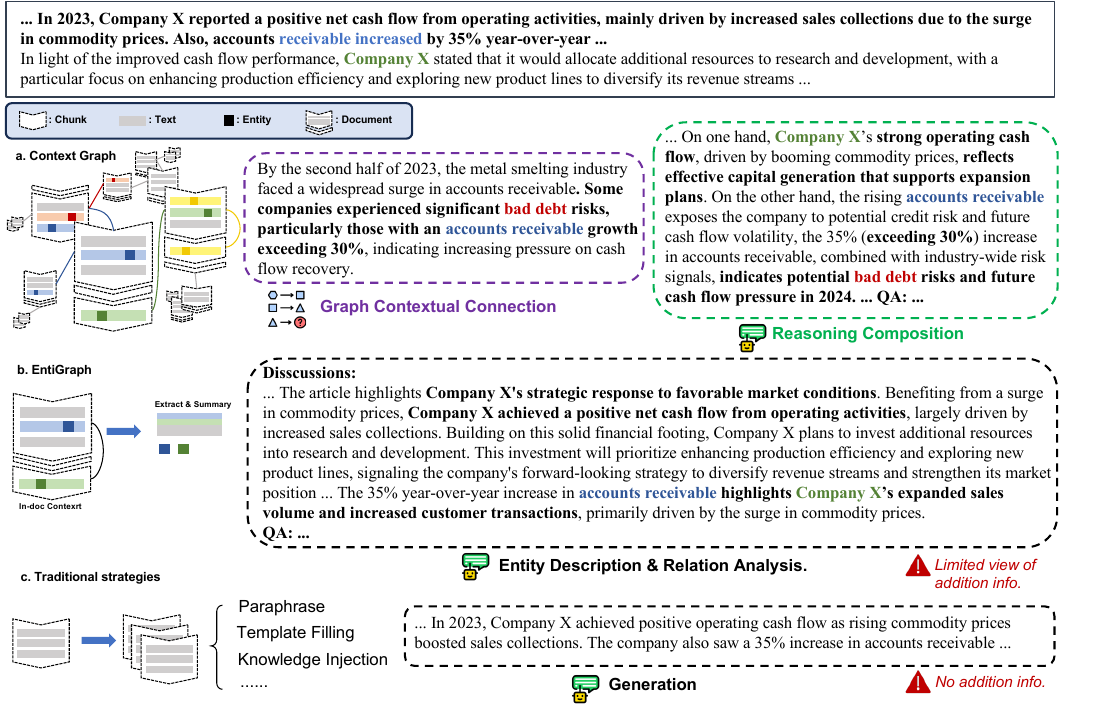}
\end{center}
\caption{Comparison of the Proposed Context Graph for Synthetic Generation with Other Generation Strategies: a. Context Graph in SoG. b. Intra-document graph in EntiGraph, where the knowledge view is confined within a single document. c. Traditional synthetic generation methods, which struggle to incorporate extra knowledge.}
\label{context_graph}
\end{figure}

To address the challenge of efficiently acquiring knowledge from small-scale corpora, synthetic data generation methods have been proposed for continued pretraining of models. They aim to expand the original limited data by generating diverse synthetic corpora, thereby improving the learning efficiency and performance of the models. For instance, the EntiGraph method decomposes the text corpus into a list of entities and generates descriptions about the relationships between entities, attempting to populate the underlying knowledge graph of the corpus \citep{EntiGraph2024}. However, as shown in Figure~\ref{context_graph}b this approach primarily focuses on intra-document content, neglecting inter-document knowledge associations. This leads to limitations in the content diversity and knowledge depth of the synthetic data.
In reality, knowledge is often interconnected across documents and domains. Relying solely on entity combinations within a single document fails to capture the full spectrum of knowledge. Additionally, the lack of cross-document synthetic data constrains the model's ability to handle complex, multi-hop problems that require integrating information from multiple documents to derive an answer. For instance, in the context graph in Figure~\ref{context_graph}a, the first encountered literature primarily describes Company X's positive financial report and active market plans in 2023. However, relying on the across-document information associated with the entity "accounts receivable" —"companies with accounts receivable growth exceeding 30\% face a special risk of bad debt" —we can derive a broader understanding of the literature: despite the positive net cash flow, people are suggested to be particularly cautious about the potential bad debt risk associated with Company X's 35\% accounts receivable growth. Cross-document information can integrate multi-dimensional perspectives on a topic (both positive and negative), build a progressive chain of information, and uncover implicit phenomena — integrating knowledge in a way that uncovers more than what each document alone can offer, where "1+1>2".

To this end, we propose the Synthesize-on-Graph (SoG) framework—a context-graph-enhanced synthetic data generation method designed to provide an efficient solution for continued pretraining of LLMs. The core idea of SoG is to incorporate cross-document knowledge associations by constructing and leveraging a context graph to expand the original corpus effectively.

Specifically, SoG comprises two key components: (1) \textit{Context Graph Construction and Cross-Document Sampling}: We build a context graph from entities and concepts extracted from the original corpus, representing cross-document knowledge associations. Using this graph, we apply a two-stage cross-document sampling strategy: first, random walks guided by document retrieval to achieve cross-document exploration, enhancing data diversity while preserving coherence and knowledge associations. This helps the model learn complex knowledge structures, especially for long-tail entities. Second, Secondary Sampling and Controlled Allocation help balance the knowledge distribution and support flexible data customization. (2)\textit{Combined Chain-of-Thought and Contrastive Clarifying Synthesize}: We combine Chain-of-Thought and Contrastive Clarifying to enhance synthetic data quality. CoT guides the model to generate logical chains, improving depth and interpretability, while contrastive generation boosts the discriminative knowledge in the synthetic data.

Through extensive experiments, our approach outperforms existing state-of-the-art (SOTA) methods on multi-hop document and domain-specific question answering tasks, while achieving comparable results on long-context reading comprehension.
 Also, we demonstrate better generalization capability over the SOTA method.
The introduction of the SoG framework marks a significant advancement in synthetic data generation and continued pertaining (CPT) for LLMs, providing new directions and possibilities for future research. Our work not only drives the development of synthetic data techniques but also offers new perspectives for optimizing the training of LLMs.

\section{Related Work}
\label{sec:related-work}
This section presents an overview of recent developments in synthetic data generation for the pretraining of large language models (LLMs). Synthetic data generation has emerged as a crucial area of research, with various strategies proposed to enhance the diversity and effectiveness of training datasets. A significant trend in this domain is the adoption of hierarchical prompting to generate targeted synthetic content. For instance, \cite{eldan2023tinystories} utilize API-based LLMs to create children’s stories driven by specific keywords, illustrating that even smaller language models can yield fluent narratives when pre-trained on such datasets. \cite{ma2024leveraginglargelanguagemodels} achieve automatic analysis and annotation on complex data in the legal domain by using a modular multi-process pipeline, along with the injection of expert knowledge in the form of few-shot learning into each submodule. This approach was used for both pretraining and fine-tuning. This underscores the potential of hierarchical prompting in producing effective and relevant training data.

In another vein, \cite{gunasekar2023textbooks} generate diverse educational content, such as textbooks and coding exercises, by conditioning on attributes like topic, audience, and function names. The datasets generated from this method have supported the development of robust LLMs, as further explored in subsequent studies \citep{li2023textbooks, javaheripi2023phi}. However, these approaches are often hindered by a lack of public accessibility to the datasets and prompt strategies, limiting reproducibility and broader community progress. Similarly, \cite{maini2024rephrasing} focus on rephrasing existing documents to generate new training data, reporting enhancements in training efficiency through these modified versions.

While these efforts have significantly advanced the field, they primarily focus on generating intra-document content, thereby overlooking the importance of cross-document knowledge associations. This oversight limits the diversity and depth of the synthetic content, which is crucial for developing LLMs capable of understanding and integrating complex knowledge structures. The prevailing focus on intra-document generation underscores the need for novel methodologies that can address these gaps by synthesizing data that not only maintains coherence but also captures broader, interconnected knowledge domains.

Current efforts \citep{mecklenburg2024injecting, yang2025longfaithenhancinglongcontextreasoning} explore synthetic QA generation for task-specific finetuning, reflecting an emerging interest in incorporating knowledge-aware strategies into data generation. Although such strategies have demonstrated benefits for specific QA tasks, their applicability remains limited for more general-purpose tasks, indicating a gap that could potentially be filled by new data generation approaches that are untethered to any particular downstream application.

Moreover, \cite{ovadia2023fine} explore continued pretraining of Llama 2 models using synthetic paraphrases of Wikipedia articles, with mixed results regarding performance improvements. This suggests limitations in relying solely on paraphrasing techniques to enhance model knowledge and underscores the need for research into more robust methods that can generate synthetic data with greater diversity and depth.

\section{Methodology}
We propose the SoG framework, a context-graph-enhanced synthetic data generation method designed to address limitations in content diversity and knowledge association found in existing approaches. The framework achieves this by leveraging cross-document, knowledge-associated sampling, enabling the integration of information across multiple sources. Additionally, it conducts a combined data synthesis approach based on Chain-of-Thought reasoning and Contrastive Clarifying analysis, which enhance generation models' ability to reason and distinguish between complex knowledge. The following sections provide a detailed overview of the SoG framework, highlighting three core components: Context Graph Construction, Cross-Document Sampling, and Generation Strategies.
\begin{equation}
\begin{aligned}
\text{Corpus}
\xrightarrow{\text{Construction}} \text{Context Graph}
\xrightarrow{\text{Context-graph Traversal}}  \text{Path Set }\mathcal{P} \\ \xrightarrow{\text{Secondary Sampling}}  \text{Balanced Path Set }\mathcal{P^*}\xrightarrow{\text{Generation Strategies}}\text{Synthetic Data}
\end{aligned}
\end{equation}

\begin{wrapfigure}{r}{0.38\textwidth}
\vspace{-18mm}
  \centering
  \includegraphics[width=0.38\textwidth]{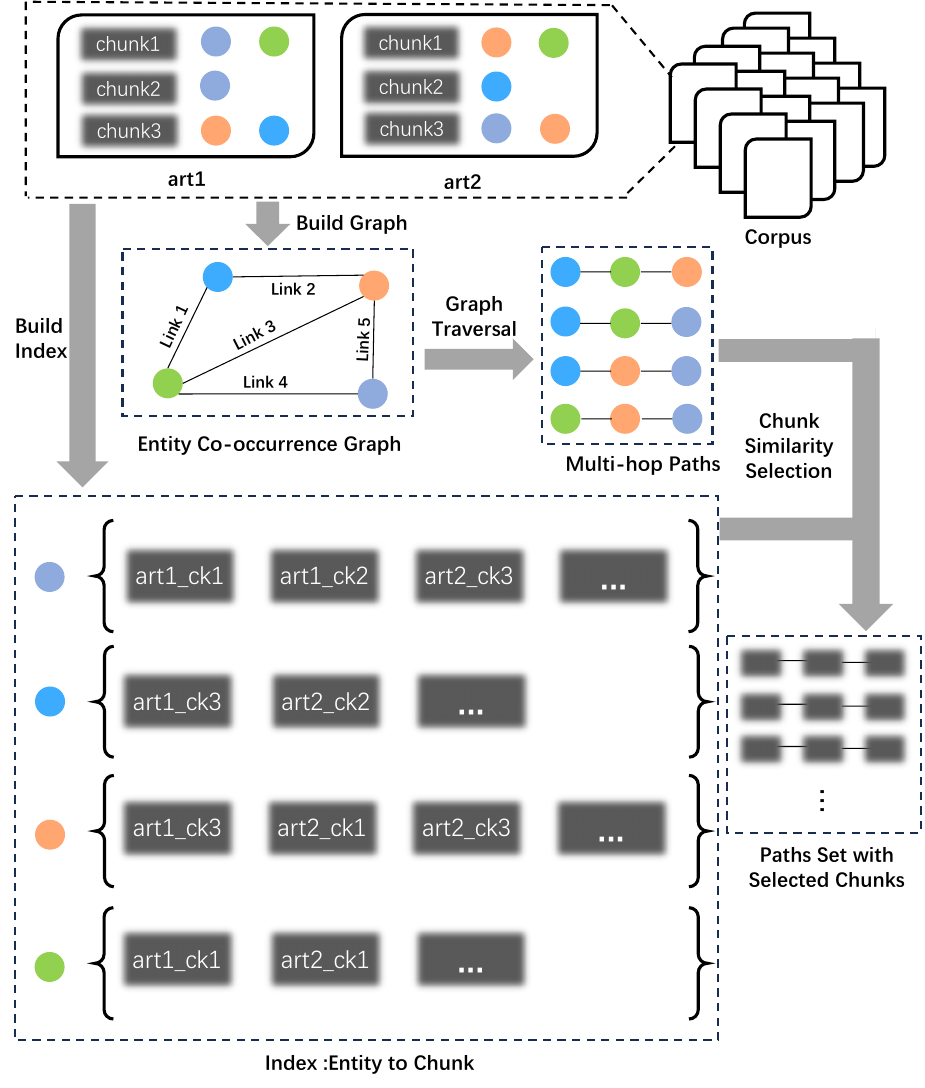} 
  \caption{Context Graph Construction and Sampling}
  \label{fig:graph_build}
\vspace{-17mm}
\end{wrapfigure}

The overall generation process and context graph building of SoG is shown in Figure~\ref{context_graph}a and Figure~\ref{fig:graph_build}.

\subsection{Context Graph Construction}
\subsubsection{Entity Extraction} 
First, given a corpus \( \mathcal{C} = \{d_i\}, i \in [0, N) \), each document \( d_i \) is divided into several paragraphs \( p_{i,j} \), where \( j \) denotes the \( j \)-th paragraph of document \( i \). Subsequently, we prompt the LLM to identify key entities within each paragraph as  \( \mathcal{E}_{i,j} \in \mathcal{E}\), where \(\mathcal{E}\) denote the extracted entities from the entire corpus $\mathcal{C}$.

\subsubsection{Entity-Context Mapping} 
For each entity \( e_k \in \mathcal{E} \), we collect all paragraphs in which it appears, denoted as \( P_k = \{p_{i,j} \mid e_k \in \mathcal{E}_{i,j}, \forall i, j\} \). This forms an entity-paragraph mapping \( M: e_k \mapsto P_k \), where \( M \) associates each entity \( e_k \) with its corresponding set of paragraphs \( P_k \).

\subsubsection{Context Graph} 
We define a context graph \( \mathcal{G} = (\mathcal{E}, E) \), where \( \mathcal{E} \) denotes the set of nodes corresponding to all identified entities. 
The edge set is given by
\[
E = \{(e_x, e_y) \mid \exists i, j \;\; \text{s.t.} \;\; e_x, e_y \in \mathcal{E}_{i,j}\},
\]
where \( \mathcal{E}_{i,j} \) represents the subset of entities co-occurring within a bounded textual unit (e.g., a paragraph or sentence). 
Thus, an edge between \( e_x \) and \( e_y \) is induced whenever the two entities are observed to co-occur within the same discourse context. 
In this way, the graph topology captures \textit{implicit contextual associations} among entities, with co-occurrence serving as a distributional proxy for semantic relatedness.

\subsection{Cross-Document Sampling}
\subsubsection{Initialization}

To enhance content diversity and knowledge association across multiple documents, we implement a cross-document sampling strategy that traverses the constructed context graph \( \mathcal{G} = (\mathcal{G}, \mathcal{R}) \). Starting from a root entity \( e_{root} \in \mathcal{E} \), we perform a breadth-first search (BFS) traversal to collect multi-hop paths that link related entities and their associated text paragraphs across documents. We will traverse all nodes in $\mathcal{G}$ as a root entity.

In addition, for each $e_{root}$, we traverse all its paragraphs using the entity-context mapping \( M \), which associates entities with the paragraphs in which they appear. If an entity occurs in a large number of paragraphs, we limit the number of starting paragraphs by randomly sampling up to \( S \), a predefined hyperparameter. This step is crucial, as the selected starting paragraphs \( p^{(0)} \) serve as references for computing embedding similarities during the traversal process.

Briefly, each entity serves as the root $e_{root}$. Then a graph traversal is performed up to a maximum of \( S \) steps according to the number of paragraphs $P_{root}$ from the mapping $M$.

\subsubsection{Context-graph Traversal}
At each traversal, step up to a specified depth \( D \), we explore neighboring entities of the current entity \( e \). The neighbors are defined as:
\[
N(e) = \{ e' \mid (e, e') \in E \},
\]
where \( (e, e') \) indicates an edge in the context graph signifying a contextual connection between entities \( e \) and \( e' \).

To prioritize neighboring entities with relevant contexts, we introduce a similarity-based selection mechanism. For each neighboring entity \( e' \), we compute a similarity score \( F_{sim}(q^{(0)}, c) \) between the root paragraph \( q^{(0)} \) associated with \( e_{root} \) and candidate paragraphs \( c \) associated with \( e' \). The similarity function \( F_{sim}() \) can be based on semantic similarity measures such as the dot product of embeddings:
\[
F_{sim}(q^{(0)}, c) = \text{dot}(\text{embed}(q^{(0)}), \text{embed}(c)).
\]
We select the top \( W \) paragraphs with the highest scores, along with their corresponding entities, to include in the sampling paths.

After $D$ steps, every traversal results in multiple paths originating from $(e_{root}, p^{(0)})$, each path representing a sequence of contextually connected entities and their associated text paragraphs across different documents. Formally, for the root entity \( e_{root} \), we construct a set of paths \( \mathcal{P} = \{ P \} \), where each path \( P \) is defined as:
\[
P = [(e_{root}, q^{(0)}), (e_1, c_1), \dots, (e_n, c_n)], \quad n \leq d,
\]
with \( e_i \in \mathcal{E} \) and \( c_i \) being the associated paragraph of \( e_i \).

By aggregating the information from these cross-document paths, we achieve greater diversity through a richer and more varied combination of cross-document knowledge. Additionally, the paths effectively capture and reflect the implicit contextual association between knowledge elements spanning multiple documents.

\subsubsection{Secondary Sampling and Controlled Allocation}
Before proceeding to the generation phase, it is crucial to consider the utilization rate and coverage of the original corpus during generation to balance the knowledge distribution, reduce redundancy, and compensate for long-tail knowledge. Therefore, we apply \textbf{secondary sampling} on $\mathcal{P}$ to selectively collect paths for generation. Specifically, we prioritize the inclusion of paths containing entities that appear less frequently in the secondary sampled path set, by accounting for the sum of utilization rate in every path. This strategy ensures a more uniform distribution of knowledge occurrences, which mitigates biases and promotes diversity within the sampled paths, thereby enhancing the overall generation quality and efficiency. 

To further refine the control over synthetic data size, we iteratively allocate the secondary sampled paths into subsets according to the coverage of the original corpus, where each subset functions as an independent unit opted for the maximum corpus coverage ($>r$) and the most balanced paragraph frequency. This modular approach allows for seamless flexibility in data customization during the generation process: depending on the required volume of synthetic data, we can combine an appropriate number of subsets to support various experimental configurations. Specifically, due to the decreasing availability of sparse entities and associated texts as sampling iterations progress, the subset obtained in the first iteration should have the highest coverage $r$ of the original corpus. As the number of iterations increases, the coverage of subsequent subsets will gradually decrease under the fixed sampling size. We use the sample size of the first subset and corpus coverage $r$ as references and, based on the difference between the current iteration's sampling rate $\triangle r =\frac{r - r'}{r}$, re-sample and re-use texts of entities with the lowest utilization rate to complete the current path subset.

\subsection{Generation Strategies}
Given a path, we design prompts to guide the LLM in generating diverse and reliable synthetic data based on the text chunk of the entities along the path.

\subsubsection{Generation Prompt}
To produce coherent and informative content from the aggregated cross-document paths, we design two generation strategies: Chain-of-Thought (CoT) and a complementary strategy Contrastive Clarifying (CC), which are shown in Figure~\ref{cot} and Figure~\ref{cc}. 

We observed that the CoT generation method significantly improves training performance. CoT serves as a more general generation strategy, applicable to all entities with graph path connections. However, for entities with sparse graph connections—those lacking rich relationships within the graph—CoT's effectiveness can be limited, as fewer paths are available and may not provide enough context to generate comprehensive relationships with other entities.

To address this challenge, we apply CC synthetic to supplement CoT synthetic for these sparse entities. Unlike CoT, CC does not rely on graph path connections, enabling it to work effectively even with entities that have limited graph relationships. Specifically, in the secondary sampling process mentioned before, we continuously monitor the current corpus coverage rate $r'$. When the total number of samples exceeds a hyperparameter $l$ and $r'$ does not reach $r$, CC is triggered for the $\triangle r$ least sampled entities based on their utilization rate. CC will randomly pair these entities without replacement. If there are $N$ least sampled entities, then the $N/2$ path will be built for CC generations.
By doing so, we enrich the generation process, helping balance the model bias caused by the long-tail distribution of entities. Furthermore, CC can explicitly clarify the differences and similarities between entities in terms of their attributes and background knowledge. This can improve the model’s discriminative power of sparse entities, providing deeper insights into their nuances. 


\textbf{CoT generation:} We prompt the LLM to fully utilize the key information from each text fragment and build a step-by-step narrative where each text fragment logically leads to the next, forming a clear flow of cause and effect. The primary goal is to synthesize information from various sources into a logically connected storyline, which ensures that the generated content is coherent and that the relationships among the fragments are explicitly articulated. 

Specifically, the narrative is structured into distinct phases—including initiation, development, turning points, and conclusion—with natural transitions that preserve the logical flow of causal relationships. Based on the constructed narrative, we prompt the LLM to formulate questions that require an understanding of the entire information chain to answer. The answers are provided in a chain-of-thought style, breaking down the reasoning process step by step to arrive at the final conclusion. This design can improve interpretability and provide deeper insight into the synthetic content.

\textbf{Contrastive Clarifying:} We prompt the LLM to generate a comparative analysis that contrasts and compares multiple text fragments. This approach is designed to prompt the LLM to explicitly analyze and highlight the implicit nuances or lack of direct connections between pieces of information, ensuring that such contrasts are clearly reflected in the synthetic data. By conducting a detailed comparative analysis, the model can effectively uncover and present discriminative information, enriching the groundedness and diversity of the synthetic content.

Specifically, the LLM is instructed to examine each entity or fragment individually, synthesize a thoughtful contrastive narrative, and summarize the comparative insights in a concluding section.  When direct similarities are absent, the narrative shifts to highlighting the unique contributions or perspectives that each entity offers within its respective context. The generated output maintains an objective and analytical tone, avoiding any attempt to force connections between unrelated fragments.


\section{Experiments}
To comprehensively evaluate the effectiveness and applicability of the proposed Synthesize-on-Graph (SoG) framework, this section explores its performance through a series of carefully designed experiments. The experiments aim to assess SoG's contributions in four major aspects: First, to what extent does incorporating cross-document knowledge associations in SoG enhance the diversity and depth of synthetic data compared to intra-document-focused methods (\textbf{RQ1})? Second, does SoG's synthetic data provide consistent performance gains across language models of different sizes (\textbf{RQ2})? Third, to what extent can SoG mitigate the long-tail knowledge problem in the original corpus (\textbf{RQ3})? In what scenarios is SoG synthesis applicable? (\textbf{RQ4})?

\subsection{Datasets}
To address our research questions, we evaluate on three representative datasets: \textsc{MultiHop-RAG}, \textsc{BioASQ}, and \textsc{QuALITY}. A detailed description of each dataset is provided in the Appendix~\ref{a1.data}.

\subsection{\textbf{Baselines and Metrics}}
We choose Direct QA (directly answering by the base model), Rephrasing (back‑translation and synonym replacement, following \cite{wei2019edaeasydataaugmentation}) and the state-of-the-art methods EntiGraph~\cite{EntiGraph2024} as baselines for evaluation. The evaluation metrics for \textsc{MHRAG}, \textsc{BioASQ} and \textsc{QuALITY} are Exact Match (EM), model-based
evaluation (MBE) approach using LLM-as-a-Judge\cite{gu2025surveyllmasajudge}, and Accuracy (Acc), respectively.

\subsection{Experiment Details}
In our generation setup, we used GPT-4o-mini as the generation model. The temperature was set at 0.7. We utilize semantic chunking\footnote{\url{https://python.langchain.com/docs/how_to/semantic-chunker/}} to split the long contexts. The semantic embedding was computed by bge-small-en-v1.5. In all experiments, we continued pretrain the LLMs with a context length of 2048 and a batch size of 64. We apply a linear learning rate warmup for 10\% of the total steps, followed by a cosine decay with a peak learning rate of 5e-6. We perform full-parameter training for 2 epochs in BF16 precision, using a per-device batch size of 2 and accumulating gradients over 4 steps. In addition, within $4.5\times$ of the original corpus size, the sampling paths for CoT are of one-hop length, while beyond that, the sampling paths are of two-hop length. For \textsc{QuALITY}, we followed the evaluation setup in EntiGraph. For \textsc{MHRAG}, we evaluate the CPT models with zero-shot prompting on a sample of 1,000 QA pairs. For \textsc{BioASQ}, we constructed a hard subset consisting of 1,114 questions that Qwen3-8B failed to answer correctly in a single attempt. This sampling criterion ensures that the selected questions reflect genuine challenges for strong LLMs, thereby providing a more rigorous evaluation of knowledge-intensive reasoning. For entity ambiguity issue, we rely on surface-form string matching combined with simple heuristics, including normalization of singular/plural forms and letter casing, alias matching, and Wikipedia-style redirect mappings to partially address this issue.


\subsection{Main Experiment Results}

\begin{wrapfigure}{r}{0.5\textwidth}
\vspace{-4mm}
\centering
\begin{tikzpicture}
\begin{axis}[
    width=0.5\textwidth,
    xlabel={Synthetic Size (Multiples of Original Corpus)},
    ylabel={Score (\%)},
    xtick={0,1,2,3,4,5,6,7},
    xticklabels={0, 0.7, 1.5, 3, 4.5, 6, 9, 23},
    ytick={10,20,30,40,50,60,70,80},
    xmin=0, xmax=7.5,
    ymin=10, ymax=80,
    legend style={
        at={(0.5,-0.5)},
        anchor=north,
        column sep=1ex,
        legend columns=2,   
        font=\scriptsize,
        cells={anchor=west}
    },
    grid=major,
    axis lines=left,
]

\addplot[blue, mark=*, semithick, mark options={fill=blue!20!white}] coordinates {
    (0, 55.3) (1, 66.3) (2, 70.9) (3, 73.2) (4, 74.1) (5, 75.4) (6, 76.1) (7, 76.3)
};
\addlegendentry{SoG (\textsc{MHRAG})}

\addplot[cyan!60!blue, mark=o, semithick, mark options={fill=cyan!20!white}] coordinates {
    (0, 37.4) (1, 42.1) (2, 44.0) (3, 46.7) (4, 46.2) (5, 47.3) (6, 48.9) (7, 50.9)
};
\addlegendentry{SoG (\textsc{QuALITY})}

\addplot[purple, mark=*, semithick, mark options={fill=purple!20!white}] coordinates {
    (0, 13.5) (1, 18.2) (2, 20.5) (3, 27.2) (4, 31.0) (5, 37.1) (6, 36.4) (7, 39.2)
};
\addlegendentry{SoG (\textsc{BioASQ})}

\addplot[red, mark=square*, semithick, mark options={fill=red!20!white}] coordinates {
    (0, 55.3) (1, 57.7) (2, 59.0) (3, 54.9) (4, 55.2) (5, 55.5) (6, 54.5) (7, 57.8)
};
\addlegendentry{EntiGraph (\textsc{MHRAG})}

\addplot[orange, mark=triangle*, semithick, mark options={fill=orange!20!white}] coordinates {
    (0, 37.4) (1, 43.4) (2, 44.8) (3, 46.2) (4, 47.4) (5, 47.8) (6, 48.3) (7, 52.1)
};
\addlegendentry{EntiGraph (\textsc{QuALITY})}

\addplot[magenta, mark=square*, semithick, mark options={fill=magenta!20!white}] coordinates {
    (0, 13.5) (1, 15.6) (2, 19.3) (3, 20.8) (4, 25.5) (5, 24.2) (6, 29.6) (7, 28.0)
};
\addlegendentry{EntiGraph (\textsc{BioASQ})}

\addplot[gray!60!black, mark=o, dashed, semithick, mark options={fill=gray!10!white}] coordinates {
    (0, 55.3) (2, 50.8) (4, 53.4) (6, 55.9)
};
\addlegendentry{Rephrasing (\textsc{MHRAG})}

\addplot[gray!30!black, mark=triangle*, dashed, semithick, mark options={fill=gray!10!white}] coordinates {
    (0, 37.4) (2, 37.9) (4, 39.1) (6, 38.7)
};
\addlegendentry{Rephrasing (\textsc{QuALITY})}

\addplot[gray, mark=diamond*, dashed, semithick, mark options={fill=gray!20!white}] coordinates {
    (0, 13.5) (1, 12.9) (2, 14.3) (3, 15.6) (4, 17.8) (5, 16.5) (6, 16.6)
};
\addlegendentry{Rephrasing (\textsc{BioASQ})}

\addplot[black, thick] coordinates {(0, 52.95) (7.5, 52.95)};
\addlegendentry{Raw CPT (\textsc{MHRAG})}

\addplot[black, thick, dashed] coordinates {(0, 35.2) (7.5, 35.2)};
\addlegendentry{Raw CPT (\textsc{QuALITY})}

\addplot[black!60, thick, dotted] coordinates {(0, 14.1) (7.5, 14.1)};
\addlegendentry{Raw CPT (\textsc{BioASQ})}

\end{axis}
\end{tikzpicture}
\caption{Performance trends of SoG and EntiGraph across three benchmarks with rephrasing and CPT baselines.}
\label{fig:performance_trends}
\vspace{-6mm}
\end{wrapfigure}
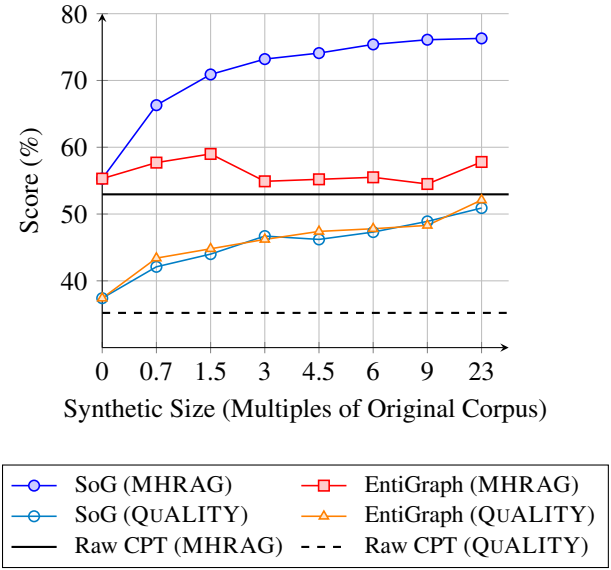

To answer \textbf{RQ1} and \textbf{RQ2}, we compare the effectiveness of SoG, traditional Rephrasing augmentation and the intra-document-focused method EntiGraph in continued pre-training (CPT) with varying amounts of synthetic data on two datasets. The results are shown in Figure~\ref{fig:performance_trends}. For \textsc{MHRAG} and \textsc{BioASQ}, model performance steadily improves as the amount of SoG synthetic data increases. In contrast, EntiGraph synthetic data provides limited gains. Especially in \textsc{MHRAG}, when the EntiGraph data size exceeds 1.5 times the original corpus, performance plateaus or even degrades due to its reliance on intra-document associations. This limitation prevents diverse and deeper generations, especially for complex tasks requiring cross-source knowledge integration.
The sharp performance gap on \textsc{MHRAG} underscores the strength of SoG’s cross-document knowledge integration in the context graph, which uncovers implicit entity relationships and enables richer reasoning. In addition, the most significant performance boost from SoG occurs when the synthetic data volume is within 0 to 1.5 times the original corpus, demonstrating that even a moderate amount of SoG data effectively enhances large model performance.

Although SoG exhibits slightly weaker performance on the \textsc{QuALITY} dataset, its results remain largely comparable EntiGraph. This modest decline stems primarily from SoG’s design emphasis on flexibility and generalizability across tasks that rely on large, interconnected corpora. In contrast, \textsc{QuALITY} poses a distinct challenge: each document is an independent narrative with minimal shared knowledge or cross-document links. To better align with this task, we constrained SoG’s path sampling strategy to operate strictly within individual documents. To align with this characteristic, we constrained SoG’s sampling strictly within individual documents. Despite that SoG's core strength, cross-document knowledge aggregation, was not fully utilized on this dataset, it still performed comparably with the SOTA method. This underscores the better generalization capability of our SoG.

Moreover, the traditional Rephrasing augmentation method yields only marginal or negligible improvements across all datasets, further highlighting the necessity of structurally informed synthetic data construction.

Finally, another observation is that CPT solely on the original corpus yields at best limited gains and in some cases even degrades performance relative to the original model (see Raw CPT in Figure~\ref{fig:performance_trends}). We attribute this to the lack of diversity and distributional differences in the original corpus, which further emphasizes the critical role of Synthetic CPT.

\subsection{Ablation Study}
\subsubsection{Influence Over Different Generation Strategy}\label{4.5.1}
\textbf{Distribution of Synthetic Data of Different Generation Strategy (\textbf{RQ3}):} The long-tail issue of entities in the original corpus may result in insufficient learning, thereby affecting the model's performance and accuracy. Additionally, the long-tail problem can cause the model to over-rely on high-frequency entities and further diminish its ability to recognize and understand rare entities. To investigate whether SoG synthetic data can alleviate the long-tail problem of entities in the original documents, we analyzed the entity distributions in the original corpus and in SoG synthetic corpora of varying sizes. 

\begin{figure}[h]
\centering
\subfloat[Entity distribution: the original corpus (1.5M).]{
    \includegraphics[width=0.3\textwidth]{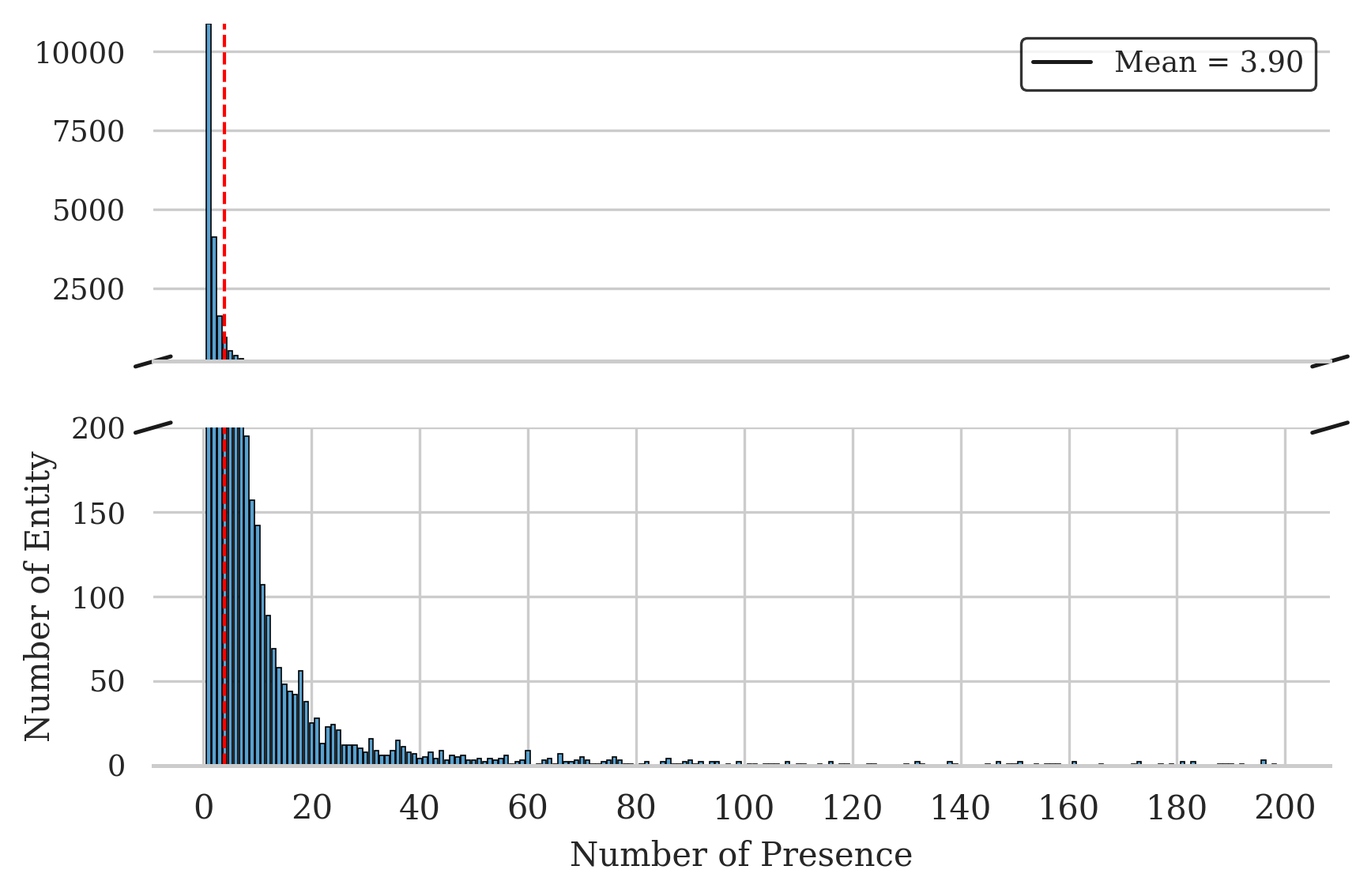}
    \label{dist2}
}
\hfill
\subfloat[Entity distribution: 6M SoG synthetic data with CoT generation.]{
    \includegraphics[width=0.3\textwidth]{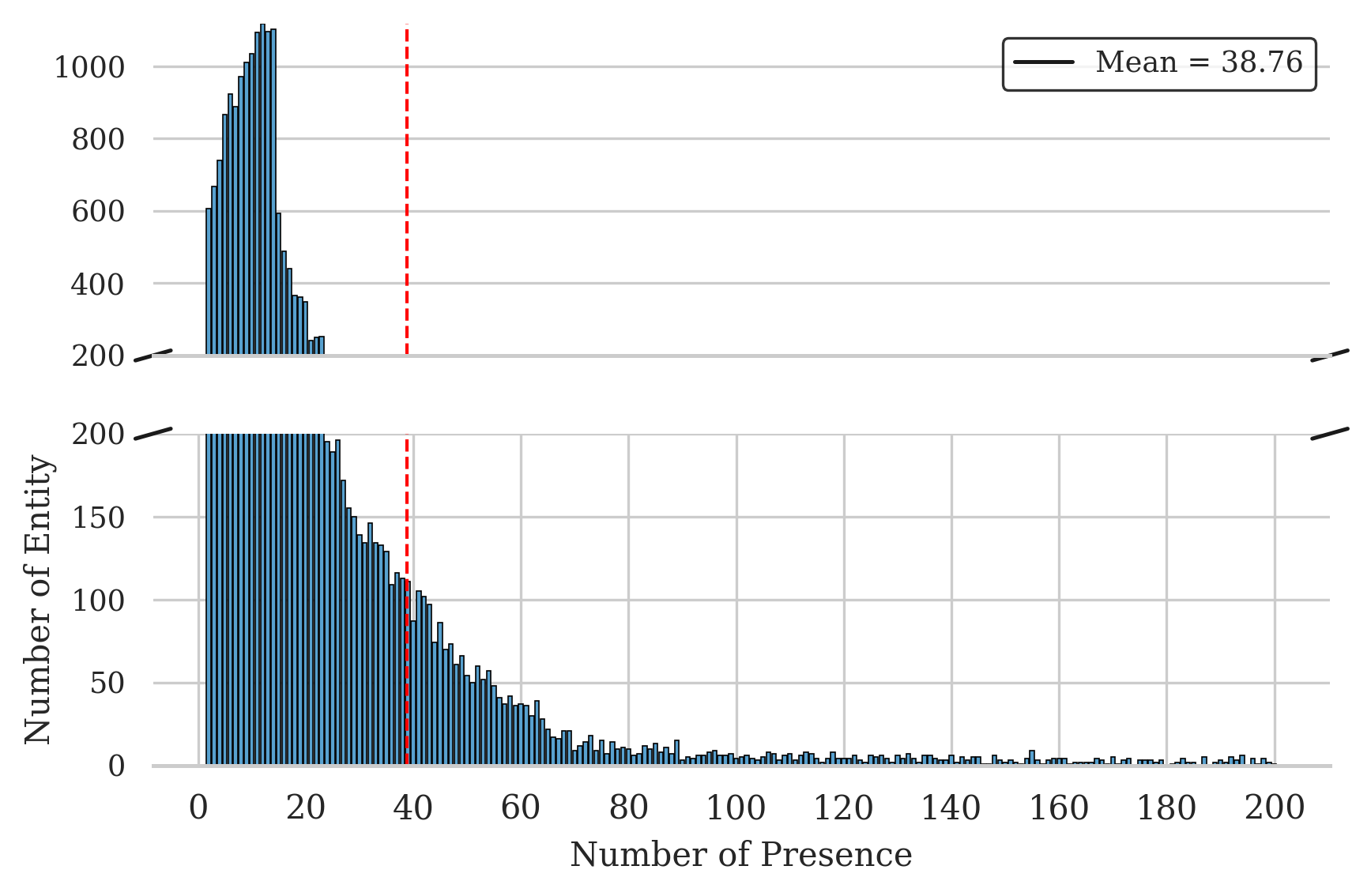}
    \label{dist1}
}
\hfill
\subfloat[Entity distribution: 6M SoG synthetic data with CoT and CC generation.]{
    \includegraphics[width=0.3\textwidth]{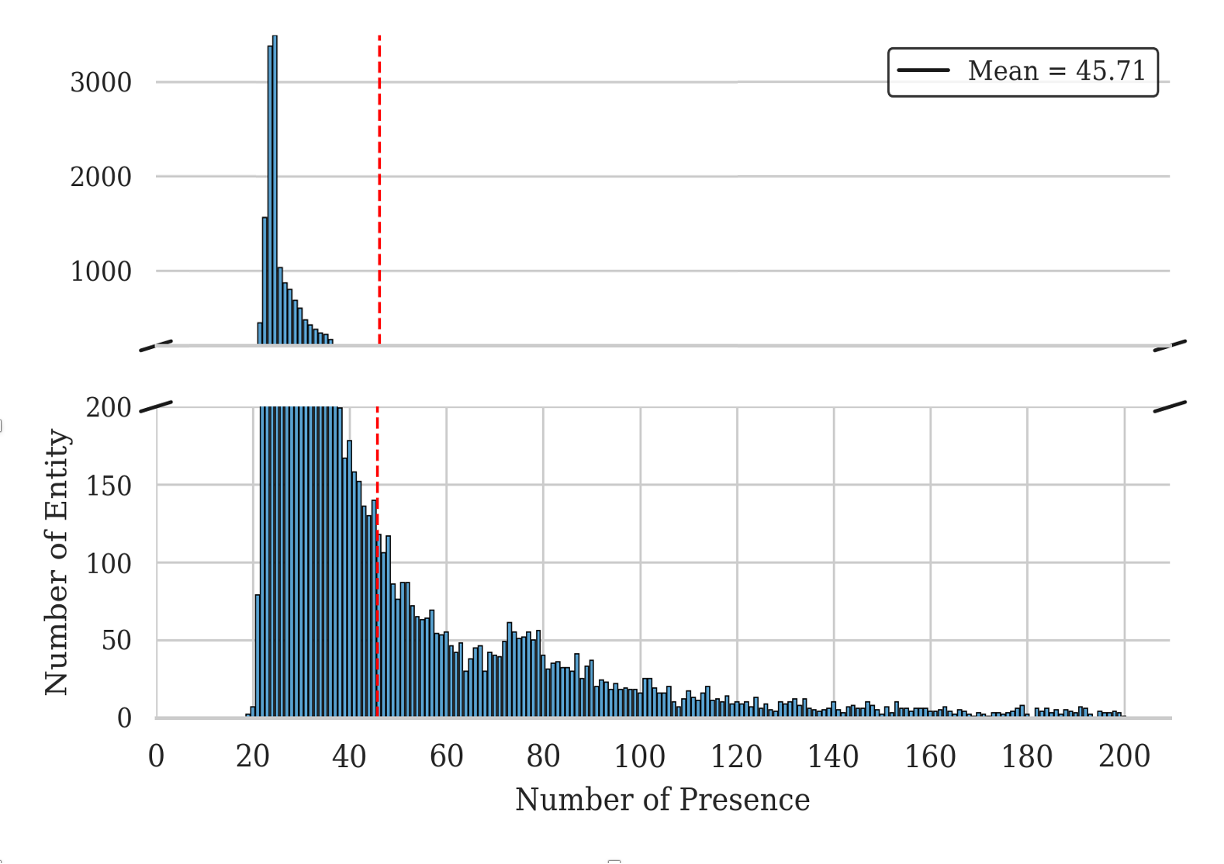}
    \label{dist3}
}
\caption{Entity distributions for different data sets.}
\end{figure}

As illustrated in Figure \ref{dist1}, \ref{dist2} and \ref{dist3}, entities in the original corpus exhibit a significant long-tail distribution. In the sampling process using only the CoT strategy (which selects paths by prioritizing entities with the lowest occurrence counts), the overall distribution becomes more concentrated. However, the long-tail trend still remains. When the Contrastive Clarifying (CC) strategy is introduced to supplement CoT (periodically enhancing long-tail knowledge based on sampling utilization rates), all long-tail entities are adequately covered, and the overall distribution begins to approximate a normal distribution. This significantly alleviates the issue of insufficient occurrences for most entities and improves diversity, demonstrating that our SoG framework can effectively balance the distribution of synthetic data.

\textbf{Training Performance of Different Generation Strategy:} 
CC is designed to specifically enhance the LLM's understanding of long-tail entities and is not suitable for standalone application to the entire corpus. As a result, synthetic data solely through CC tends to be of lower quality compared to that produced by CoT. CoT primarily focuses on generating additional useful information by integrating knowledge across documents. Therefore, CoT alone can already achieve sufficient synthetic data quality. However, due to their low frequency, long-tail entities often receive less attention from CoT. As shown in the \textsc{MHRAG} results in Table~\ref{tab:performance_comparison}, combining both generation strategies can further improves the effectiveness of synthetic data for CPT training. Interestingly, on the \textsc{QuALITY} dataset, using CoT alone outperforms the combined strategy. We believe this is because each QA pair in \textsc{QuALITY} is based on a single novel and does not involve cross-document knowledge. Such tasks tend to focus less on long-tail entities and more on the main plots and characters within the document. In this case, the CoT strategy naturally aligns with the primary content of the story. For different scenarios, our approach allows flexible adjustment of the sampling and synthesis strategies in SoG to better align with the feature of the original corpus and the specific task requirements. The specific SoG configuration adjustments for \textsc{QuALITY} are provided in the Appendix~\ref{aqlt}.
\begin{table*}[ht]
\centering
\begin{minipage}{0.55\linewidth}
\caption{Performance of Different Approaches on Llama-3-8B-Instruct}
\label{tab:performance_comparison}
\centering
\begin{adjustbox}{width=\linewidth}
\begin{tabular}{lcccc}
\toprule
\textbf{Dataset} & \textbf{CoT + CC} & \textbf{CoT} & \textbf{CC} & \textbf{Direct QA} \\
\midrule
\textsc{MHRAG}(X$1.5$) & $70.9$ & $70.6$ & $63.7$ & $55.3$ \\
\textsc{MHRAG}(X$4.5$) & $74.1$ & $72.9$ & $62.6$ & $55.3$ \\
\textsc{QuALITY}(X$1.5$) & $44.0$ & $44.7$ & $38.9$ & $37.4$ \\
\textsc{QuALITY}(X$4.5$) & $46.2$ & $47.5$ & $42.8$ & $37.4$ \\
\bottomrule
\end{tabular}
\end{adjustbox}
\end{minipage}
\hspace{5mm}
\begin{minipage}{0.35\linewidth}
\caption{CPT vs. RAG results: \textbf{Base LLM} denotes Llama-3-8B-Instruct. \textbf{CPT LLM} denotes the model CPT on the SoG data. \textbf{Zero-shot} denotes directly answering by the corresponding model.}
\label{tab:rag}
\centering
\begin{adjustbox}{width=\linewidth}
\begin{tabular}{lcc}
\toprule
\textbf{Model} & \textbf{RAG} & \textbf{Zero-shot} \\
\midrule
Base LLM     & $73.5$ & $55.3$ \\
CPT LLM & $70.7$ & $73.2$ \\
\bottomrule
\end{tabular}
\end{adjustbox}
\end{minipage}
\end{table*}
\subsection{CPT vs. RAG}
In this experiment, we aim to answer whether non-parametric external knowledge in retrieval-augmented generation (RAG) can be replaced by parametric knowledge acquired through SoG-based CPT. Specifically, we adopt Llama-3-8B-Instruct as the base model and evaluate its performance on the \textsc{MHRAG} task under three configurations: LLM with SoG CPT, LLM with RAG, and LLM with both SoG CPT and RAG. From the results in Table~\ref{tab:rag}, both RAG and CPT individually bring significant and similar performance gains to the LLM. Interestingly, applying RAG on top of the LLM already enhanced by synthetic CPT does not lead to further improvements. In fact, this combined setting performs worse than using either method alone. We argue that although RAG still holds a marginal advantage in performance, this advantage is outweighed by the broader benefits of synthetic CPT—including eliminating the need for retrieval, enabling shorter input windows for higher efficiency, and saving considerable computational costs in long term (\textbf{RQ4}). \textbf{Our findings highlight that incorporating SoG synthetic data into CPT enables parametric knowledge to streamline task adaptation and enhance output controllability, offering a more efficient alternative to reliance on inference-time retrieval.}


\section{Conclusion}
We propose Synthesize-on-Graph (SoG) framework, a context-graph-enhanced synthetic data generation method that effectively incorporates cross-document knowledge associations, which combine balanced sampling with Chain-of-Thought and Contrastive Clarifying generation strategies. Experimental results show that SoG achieves SOTA performance on multi-hop QA tasks while showing better generalization capability. Our work highlights the potential of SoG as a scalable and efficient solution for continued pretraining, offering new directions for optimizing large language model training in knowledge-intensive domains.

\bibliographystyle{unsrtnat}
\bibliography{reference}

\appendix
\section{Appendix}\label{sec:appendix}
\subsection{Datasets}\label{a1.data}

\begin{itemize}
    \item MultiHop-RAG (\textsc{MHRAG})~\cite{tang2024multihopragbenchmarkingretrievalaugmentedgeneration} is specifically designed to challenge the multi-hop reasoning capabilities of LLMs. It consists of queries constructed from news articles published between September and December 2023, which include information beyond the training cutoff of existing LLMs, ensuring that synthetic data is required to fill knowledge gaps. In addition, each query requires models to integrate evidence from multiple documents, mimicking real-world scenarios where knowledge is dispersed across sources. Existing LLMs, even RAG systems, often struggle with such tasks, underperforming in tasks that demand integrating and reasoning over scattered evidence. This dataset serves as an ideal benchmark to evaluate how SoG-generated synthetic data equips LLMs to utilize their internal knowledge for handling complex multi-hop reasoning effectively.
    \item \textsc{BioASQ}~\cite{krithara2023bioasq}: The \textsc{BioASQ} question answering (QA) benchmark dataset contains questions in English, along with golden standard (reference) answers and related material. The dataset has been designed to reflect real information needs of biomedical experts, assess the comprehensive understanding of professional knowledge, and is therefore more realistic and challenging than most existing datasets. We aim to explore challenging problems in professional domains that require highly specialized expertise, and investigate to what extent SoG can provide models with better learning corpora. 
    \item \textsc{QuALITY}~\cite{pang-etal-2022-quality} is a multiple-choice question-answering dataset for long document comprehension. Unlike in prior work with passages, the questions are written and validated by contributors who have read the entire passage, rather than relying on summaries or excerpts. For a fair comparison with the state-of-the-art CPT synthetic data method, EntiGraph, we also chose this dataset for evaluation.
\end{itemize}
\subsection{Performance on More Backbone Models}
Evaluating across more base models is crucial for assessing the robustness and generalizability of SoG. To this end, we have conducted additional experiments using \textbf{Qwen2.5-7B-Instruct} and \textbf{Qwen2.5-32B-Instruct} on the MHRAG dataset. The results, presented below, show that, with SoG CPT, smaller models tend to yield closer performance to the larger model:

\begin{table}[ht]
\centering
\renewcommand{\arraystretch}{1.2} 
\setlength{\tabcolsep}{12pt} 
\caption{Performance of SoG on the MHRAG dataset across different backbone models.}
\begin{tabular}{lccc}
\toprule
\textbf{Model} &\textbf{Direct QA} & \textbf{3$\times$} & \textbf{6$\times$} \\
\midrule
Qwen2.5-3B-Instruct & 46.7 & 67.1 \,(+43.7\%) & 73.0 \,(+56.4\%) \\
Qwen3-8B  & 50.7 & 70.5 \,(+39.0\%) & 76.4 \,(+50.7\%) \\
LLaMA-3-8B-Instruct  & 48.7 & 70.9 \,(+45.6\%) & 75.4 \,(+54.8\%) \\
Qwen2.5-32B-Instruct & 55.6 & 73.4 \,(+32.0\%) & 81.3 \,(+46.2\%) \\
\bottomrule
\end{tabular}
\label{tab:backbone_models}
\end{table}

\begin{table}[ht]
\centering
\renewcommand{\arraystretch}{1.2} 
\setlength{\tabcolsep}{12pt} 
\caption{Performance of SoG on the \textsc{BioASQ} dataset across different backbone models.}
\begin{tabular}{lccc}
\toprule
\textbf{Model} &\textbf{Direct QA} & \textbf{3$\times$} & \textbf{6$\times$} \\
\midrule
Qwen2.5-3B-Instruct & 11.8 & 21.7 \,(+83.9\%) & 29.4 \,(+149.2\%) \\
Qwen3-8B            & 10.3 & 20.9 \,(+103.9\%) & 28.5 \,(+176.7\%) \\
LLaMA-3-8B-Instruct & 13.5 & 26.2 \,(+94.1\%)  & 35.1 \,(+159.3\%) \\
Qwen2.5-32B-Instruct& 27.8 & 44.5 \,(+60.1\%)  & 57.3 \,(+106.1\%) \\
\bottomrule
\end{tabular}
\label{tab:backbone_models}
\end{table}

\subsection{Influence of Path Length}
We conduct a comparison to assess the impact of different sampling path length choices on the performance of CPT training in Table~\ref{a2}. The 1-hop paths can generate up to $5\times$ the data volume; therefore, only the $4.5\times$ result is reported. In general, the 1-hop setting achieves the best performance. The data synthesized from 2-hop paths also show significant performance. However, the 3-hop paths perform considerably weaker. We believe that this may be related to the inherent difficulty of the dataset's tasks. Furthermore, considering the challenges of constructing multi-hop reasoning tasks, most reasoning tasks are designed within two hops~\cite{ma2025thinkongraph20deepfaithful}.

\begin{table}[ht]
\centering
\caption{Impact of Sampling Path Length on CPT Training Performance}
\label{tab:path_length}
\begin{adjustbox}{width=0.5\columnwidth} 
\begin{tabular}{lcccc}
\toprule
\textbf{Scale} & \textbf{1-Hop} & \textbf{2-Hop} & \textbf{1+2-Hop ($1:1$)} & \textbf{3-Hop} \\
\midrule
$4.5\times$ & $74.0$ & $71.9$ & $72.5$ & $69.3$ \\
$9\times$   & -    & $73.5$ & $76.1$ & $70.7$ \\
\bottomrule
\end{tabular}
\end{adjustbox}
\label{a2}
\vspace{-3mm}
\end{table}

\subsection{Configuration Adjustment Detail for \textsc{QuALITY}}\label{aqlt}
Since each question in \textsc{QuALITY} focuses on a single article, we impose a constraint during multi-hop path sampling: All entities along the sampled path must be mapped to the same article ID to ensure that the retrieved texts come from the same article. We prioritize sampling the 1-hop paths. Additionally, during synthesis, we explicitly inform the LLM of the article title to which each input chunk belongs.

\subsection{Implementation Cost}
Our method does not rely on the strongest or most expensive LLMs. All generations are conducted with \textbf{GPT-4o-mini}, a fast and cost-efficient model (pricing: \$0.15 per 1M input tokens, \$0.08 per 1M cached input tokens, and \$0.60 per 1M output tokens). In the synthetic generation stage, the average input and output token counts per instance are approximately $1{,}700$ and $900$, respectively. Based on our experiments, expanding the corpus by $3\times$--$4.5\times$ (i.e., $\approx 2$--$3$M tokens for Enti-Graph) is already sufficient to yield substantial performance improvements. Consequently, the overall cost of SoG remains modest, making it a practical and accessible choice even under limited computational or financial resources.

\subsection{Limitations}
While our method shows promising results, several limitations remain.  First, although we conducted experimental analysis on the setting of sampling path length in \textsc{MHRAG}, this setting is task-dependent, and determining an appropriate setting for different datasets may require empirical tuning. Second, continued pretraining may introduce unstable LLM output, which requires additional training techniques~\cite{ke2023continualpretraininglanguagemodels}. 
We leave these for future work.

\subsection{Balanced Secondary Sampling}

\begin{algorithm*}
\caption{\textsc{SecondarySampling}}
\begin{algorithmic}[1]
\State \textbf{Input:} \texttt{PathSet}, target coverage rate $r$, standard length $l$, and entity to chunk index \texttt{EntityToChunk}.
\State \texttt{RemainingPaths} $\gets$ \texttt{PathSet}
\State \texttt{SampledPathsCollections} $\gets$ $\emptyset$
\State \textsc{Initialize }(\texttt{EntityUtilizationDict})\text{ with default value } 0
\While{$\mathcal{R} \neq \emptyset$}
    \State $\mathcal{P^*}$, $\mathcal{R}$, $\texttt{EntityUtilizationDict}$ $\gets$ 
    \State \hspace{1cm} $\texttt{BalancedSampling}(\mathcal{R}, r, \texttt{EntityUtilizationDict}, l, \texttt{EntityToChunk})$
    \State \textsc{Add} $\mathcal{P^*}$ to \texttt{SampledPathsCollections}
\EndWhile
\State \textsc{Save}(\texttt{SampledPathsCollections}) for synthetic generation
\end{algorithmic}
\end{algorithm*}

\begin{algorithm*}
\caption{\textsc{BalancedSampling}}
\begin{algorithmic}[1]
\State \textbf{Input:} remaining paths set $\mathcal{R} = \{P\}$, target coverage rate $r$, $\texttt{EntityUtilizationDict}$, standard length $l$, and entity to chunk index $\texttt{EntityToChunk}$.
\State \textbf{Output:} sampled paths set $\mathcal{P^*}$, $\mathcal{R}$, $\texttt{EntityUtilizationDict}$
\State $\mathcal{P^*}$ $\gets$ \{``cot'': $\emptyset$, ``cc'': $\emptyset$\}
\State $r' \gets 0$

\While{$\mathcal{R} \neq \emptyset$}
    \State $\mathcal{R} \gets \textsc{sort}(\mathcal{R}, \text{descending}, \text{by}$
    \State \hspace{1cm} $\textsc{PathUtilizationCount}(P) = \sum_{\texttt{node} \in P} \texttt{EntityUtilizationDict}[\texttt{node}])$
    
    \State \# In default, $l = \texttt{TotalNumberOfChunksInCorpus}/(\texttt{hop} + 1)$ and $r=100\%$
    \State $P' \gets \textsc{pop}(\mathcal{R})$
    \State \textsc{Add} $P'$ \text{to} $\mathcal{P^*}[\text{``cot''}]$
    \State \# Remove the path with the least node Utilization count.
    \State \textsc{Update} $\texttt{EntityUtilizationDict}$ \text{and} $r'$ \text{based on} $P'$
    \If{$r' \geq r$}
        \State \textsc{Break}
    \EndIf
    \If{$\textsc{len}(\mathcal{P^*}[\text{``cot''}]) \geq l$}
    \State $\triangle r \gets \frac{r - r'}{r}$ 
    \State $\textsc{Sort } \texttt{EntityUtilizationDict} \text{ in ascending order}$
    \State $k \gets \lfloor \triangle r \times l \rfloor$
    \State $\texttt{cut} \gets \lfloor (1-\triangle r) \times l \rfloor$
    \State \textsc{Add}$(\mathcal{P^*}[\text{``cot''}][\texttt{cut:}])\text{ back to }\mathcal{R}$
    \State  \textsc{Reverse} $\texttt{EntityUtilizationDict}$ \text{based on} $\mathcal{P^*}[\text{``cot''}][\texttt{cut:}]$
    \State $\mathcal{P^*}[\text{``cot''}] \gets \mathcal{P^*}[\text{``cot''}][0 \texttt{:cut}]$
    \State $\texttt{SparseEntities} \gets \texttt{EntityUtilizationDict}[0\texttt{:}k]$
    
    \For{each pair $(e_x, e_y) \in \textsc{SamplePairs}(\texttt{SparseEntities})$} 
        \State $c_x\gets$ $\textsc{SampleChunks}(\texttt{EntityToChunk}[e_x])$
        \State $c_y \gets$ $\textsc{SampleChunks}(\texttt{EntityToChunk}[e_y])$
        \State \# $\textsc{SamplePairs}$: Random combinations without replacement.
        \State \# $\textsc{SampleChunks}$: Random sample one chunk.
        \State \textsc{Add} $[(e_x, c_x), (e_y, c_y)]$ \text{to} $\mathcal{P^*}[\text{``cc''}]$
    \EndFor
    \State \textsc{Update} $\texttt{EntityUtilizationDict}$ \text{based on} $\mathcal{P^*}[\text{``cc''}]$
    \State \textsc{Break}
\EndIf
\EndWhile
\State \Return $\mathcal{P^*}$, $\mathcal{R}$, $\texttt{EntityUtilizationDict}$
\end{algorithmic}
\end{algorithm*}

\clearpage
\subsection{Prompt}

\begin{figure}[H]
\begin{center}
\includegraphics[width=0.8\columnwidth]{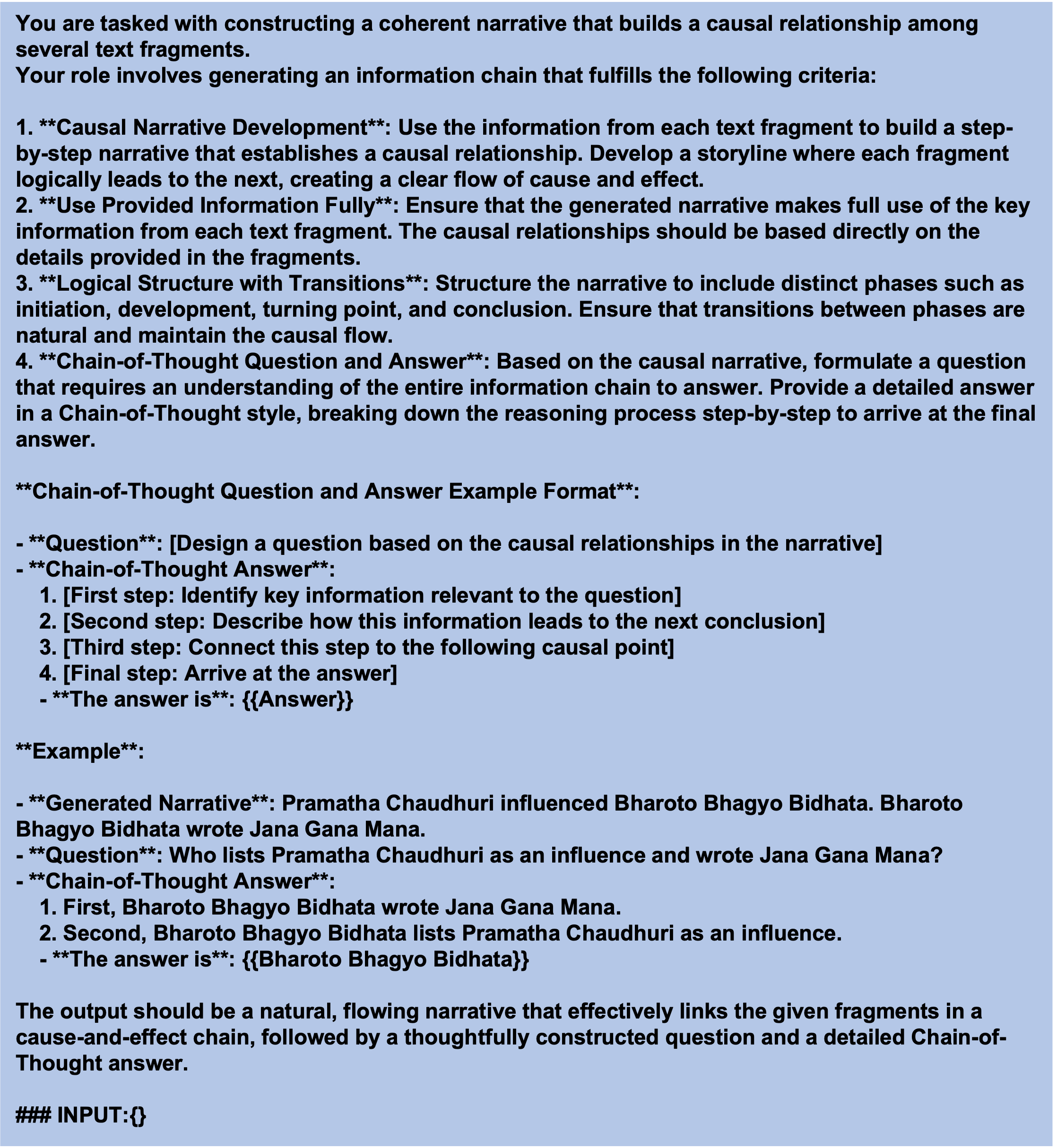}
\end{center}
\vspace{-2mm}
\caption{CoT Synthetic Prompt}
\label{cot}
\vspace{-4mm}
\end{figure}

\begin{figure}[H]
\begin{center}
\includegraphics[width=0.8\columnwidth]{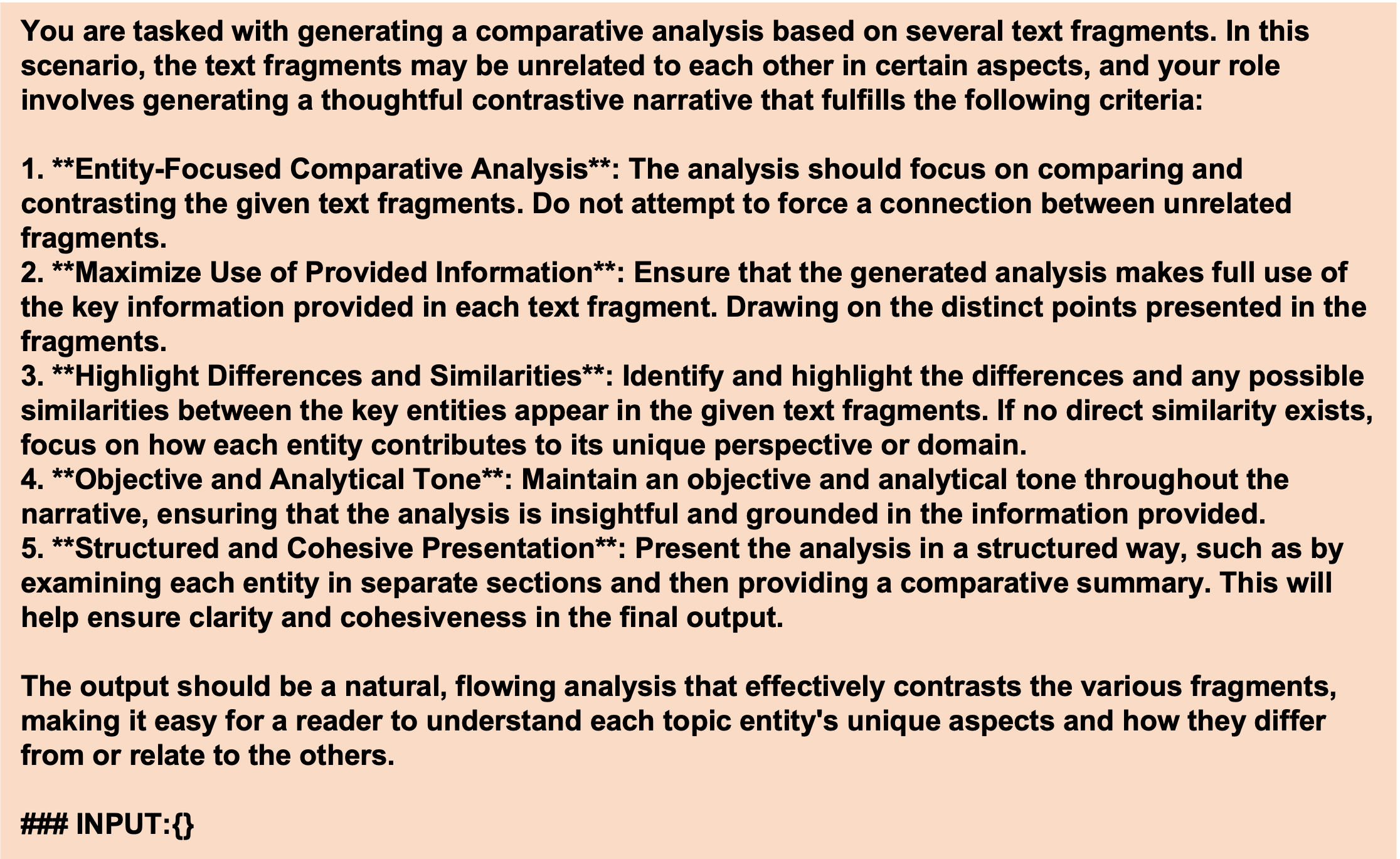}
\end{center}
\vspace{-2mm}
\caption{CC Synthetic Prompt}
\label{cc}
\vspace{-4mm}
\end{figure}

\subsection{Cases}

\begin{figure}[H]
\begin{center}
\includegraphics[width=1\columnwidth]{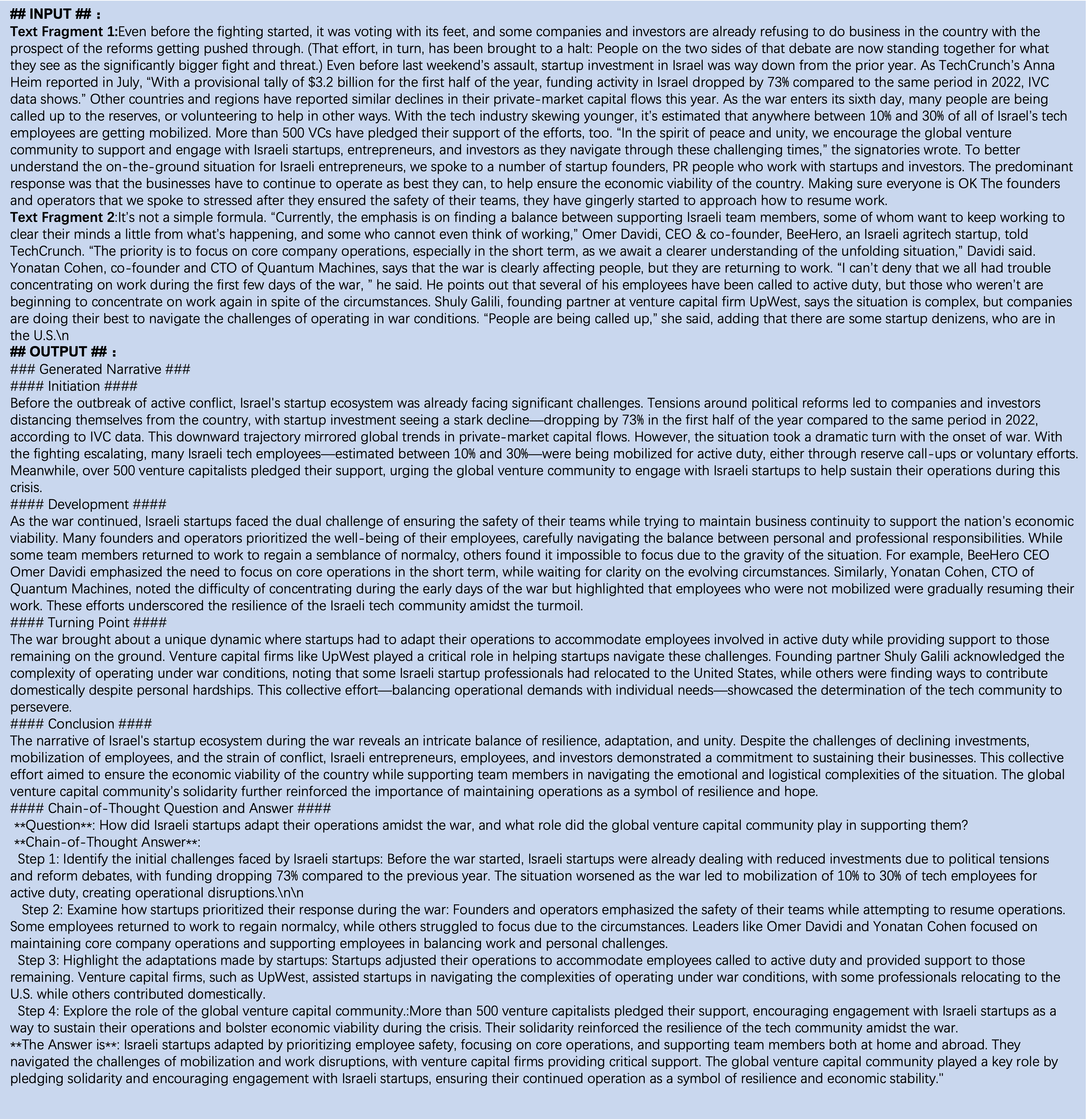}
\end{center}
\caption{CoT Case 1: 1-hop.}
\label{case1}
\vspace{-5mm}
\end{figure}

\begin{figure}[ht]
\begin{center}
\includegraphics[width=1\columnwidth]{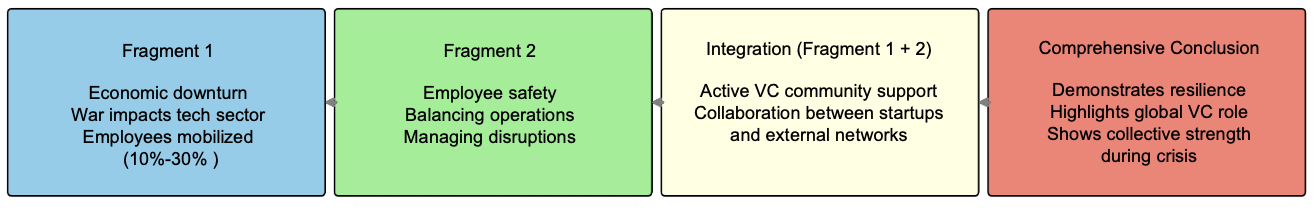}
\end{center}
\caption{Brief View: CoT Case 1}
\label{case1}
\end{figure}

\begin{figure}[H]
\begin{center}
\includegraphics[width=1\columnwidth]{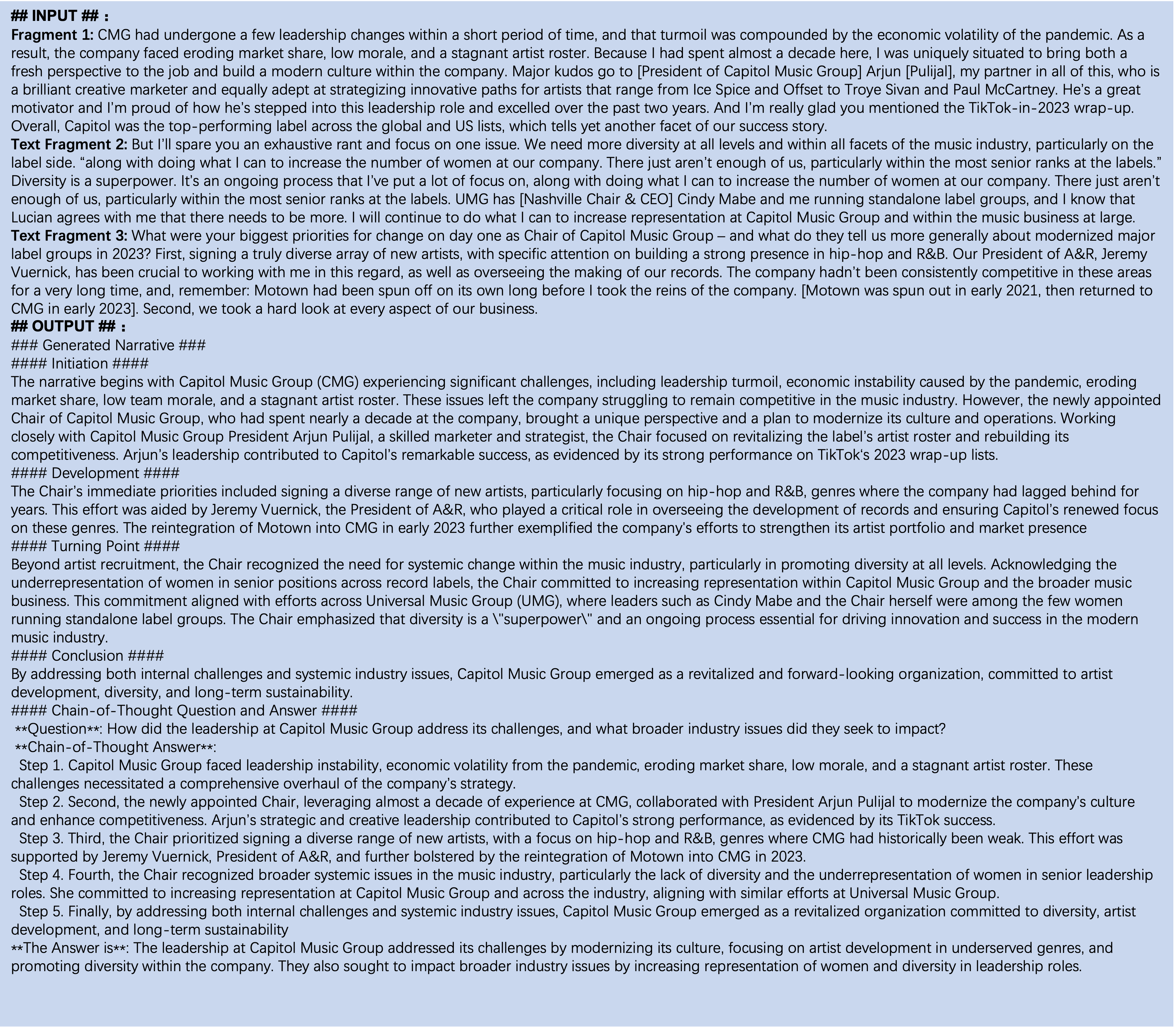}
\end{center}
\caption{CoT Case 2: 2-hop.}
\label{case1}
\vspace{-5mm}
\end{figure}

\begin{figure}[H]
\begin{center}
\includegraphics[width=1\columnwidth]{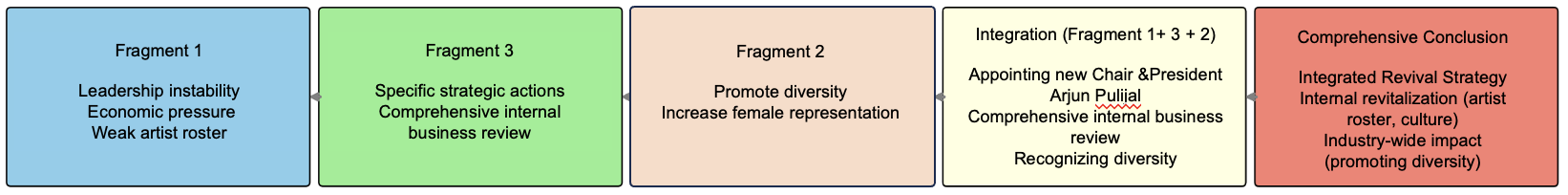}
\end{center}
\vspace{-5mm}
\caption{Brief View: CoT Case 2}
\label{case1}
\vspace{-4mm}
\end{figure}

\clearpage
\begin{figure}[H]
\begin{center}
\includegraphics[width=1\columnwidth]{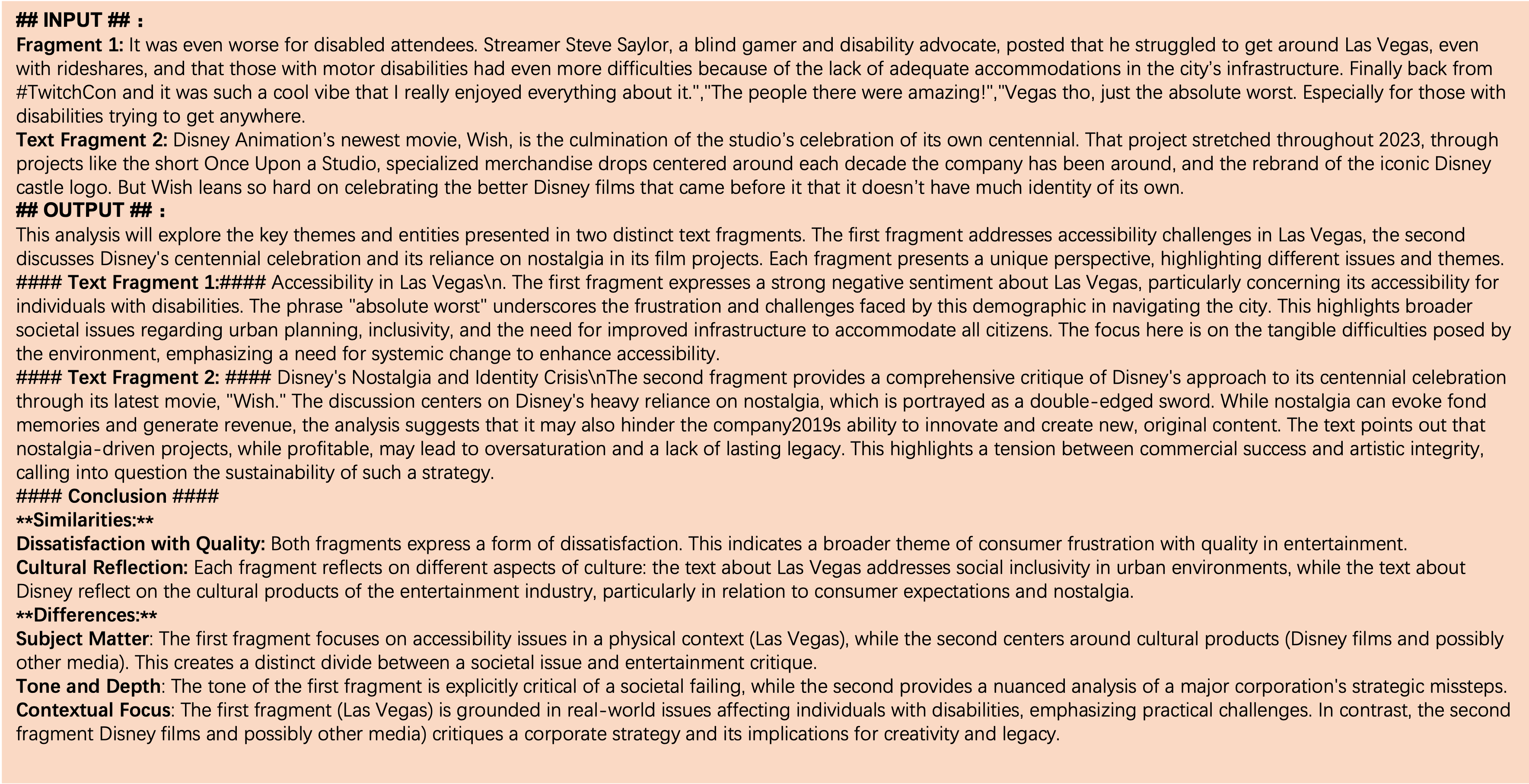}
\end{center}
\caption{CC Case: 1-hop.}
\label{case1}
\vspace{-5mm}
\end{figure}

\end{document}